# Early Prediction of Course Grades: Models and Feature Selection

Hengxuan Li
Washington University in St. Louis
h.li@wustl.edu

Collin F. Lynch
North Carolina State University
cflynch@ncsu.edu

Tiffany Barnes
North Carolina State University
tmbarnes@ncsu.edu

## ABSTRACT
In this paper, we compare predictive models for students' final performance in a blended course using a set of generic features collected from the first six weeks of class. These features were extracted from students' online homework submission logs as well as other online actions. We compare the effectiveness of 5 different ML algorithms (SVMs, Support Vector Regression, Decision Tree, Naive Bayes and K-Nearest Neighbor). We found that SVMs outperform other models and improve when compared to the baseline. This study demonstrates feasible implementations for predictive models that rely on common data from blended courses that can be used to monitor students' progress and to tailor instruction.

## 1. INTRODUCTION
In recent years, universities have begun to employ more online educational tools such as e-Textbooks, forums and homework submission systems. Online homework platforms, such as Webassign, support students and instructors by providing opportunities for automated grading and feedback. They can also support real-time monitoring of students' progress in the course.

If we can observe students' progress as they work and reliably predict their final grades, then can tailor the support provided to their needs. If, for example, a student's behavior indicates that they will succeed then simple encouragement (e.g. "keep up the good work") may be all that is required. If, however they are likely to fail, then they can be flagged for individual tutoring. Or they can be provided with automated guidance to useful resources or additional practice opportunities.

Our goal is to develop an accurate early predictor of students' final course grades from their user-system interaction logs. In order to strike a balance between early intervention and prediction accuracy. we trained our predictors based on the first 6 weeks of our 14-week course.

A number of researchers have sought to apply machine learning to predict students' course performance. Li et al. [9] proposed composite machine learning models based on features derived from students' interactions with forums, lectures, and assignments to identify at-risk students, and found that a Stacked Sparse Autoencoder+Softmax model achieved best AUC score consistently. Jiang et al. [8] sought to predict whether students would receive a completion certificate in a MOOC and if so what level it would be. To that end he combined their week 1 assignment performance with their online social interactions via logistic regression. They achieved 92.6% accuracy. Lopez et al. [10] applied a range of clustering methods to predict students' final marks in an online course based on their forum participation. They compared Expectation-Maximisation (EM) clustering, XMeans, Simple KMeans, and DTNB, using a set of four textual attributes and two network attributes: messages sent per student, replies per student, number of words written and the average expert rating of each message as well as the student's centrality and level of prestige within the social network. They found that the EM algorithm had higher accuracy than the alternatives. Similarly, Agnihotri et al. [1] applied K-Means clustering to login data from a web-based assessment platform called Connect and found a strong correlation between students' login patterns (e.g. opening assignments / attempting questions) and their scores. Brooks et al. [3] built a predictive model from time-series logs of student interactions with an online learning platform including quiz attempts, lecture views and posting to the forum. They used a decision tree to predict the students' final marks based on counting the different types of interactions over different time frames. Sabourin et al. [11] combined decision trees and Logistic Regression to classify students' self-regulated learning behaviors on an existing computer-based platform called Crystal Island. They found a weighted-by-Precision model to be most successful in classifying students' level of Self-Regulated Learning ("the process by which students activate and sustain cognitive, behaviors, and affects that are systematically directed toward the attainment of goals" [12]) performance (low, medium, high) through self-report prompts in game.

Bydzovska [5] evaluated multiple approaches to identify unsuccessful students. One approach used Support Vector machines (SVMs) and regression models based on social metrics, including measures of the students' betweenness and centrality (how many paths between students go through them). The other used collaborative filtering based on similarities between students' prior achievements. He found that the first approach reaches significantly better results for courses with a small number of students. In contrast, the second approach achieves significantly better results for mathematical courses. Stapel et al. [13] incorporated Knowledge Tracing with traditional machine learning such as K-Nearest Neighbor (KNN) and Naive Bayes to build an ensemble method to predict students' performance over specific math objectives, achieving an accuracy of 73.5%.

Holsta et al. [7] built a classification model to identify at-risk students without legacy data from other courses. This model used students' demographic, registration information in combination with online activity logs such as clicks in forum or assignment submissions. They compared the performance of these models on seven different datasets and found that XGBoost performanced better on average than SVMs, Linear Regression, KNN and Random Forests. Bote-Lorenzo et al. [2] found that Stochastic

Gradient Descent outperformed Linear Regression, SVMs and Random Forest at predicting the decrease of engagement of the students in a MOOC using a combination of assignment grades and submission statistics.

While most prior research was based on comparing the accuracy of different machine learning methods, the models used were either based on traditional onsite courses or MOOCs, and the number of features used was limited. In this paper, we built a model using generic features on homework submissions that are not unique to any specific course. We tested the accuracy of 5 different machine learning algorithms on data collected from a blended course, which pairs in-person lectures and office hours with an array of online tools including discussion forums, intelligent tutoring systems, and homework helpers. We employed Leave-One-Out cross validation to compare the accuracy of the different algorithms.

## 2. DATASET & FEATURES

We analyzed student data from CSC226 "Discrete Mathematics for Computer Scientists", a course offered by the Department of Computer Science at North Carolina State University. This is an introductory course for Computer Science (CS) and Computer Engineering students. It covers logic proofs, probability, set theory, combinatorics, graph theory, and finite automata. The dataset was collected from the Spring 2013 offering. This course has 2 lecture sections meeting 3 times per week, with 249 students total. The course lasted one semester (14 weeks) with 10 homework assignments, 2 intelligent tutors as labs and 4 tests (including the final). The final grade was based on the test scores (60%) and on the homework and lab assignments (40%). The final grade distribution is shown in Figure 1.

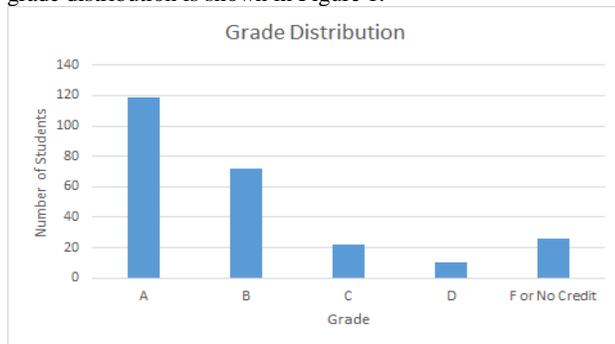

**Figure 1: The grade distribution of course analyzed**

We designed and compared a series of predictors based on the students' first 6 weeks of the coursework that includes four homework assignments and one test. The students completed their homework on Webassign, an online platform that supports automated grading and multiple retries. The homework questions were structured as short answer, fill in the blank (including Boolean values), or multiple choice questions. Complex questions such as the logic circuit shown in Figure 2, were broken into multiple submissions.

The students were typically given 1 attempt for each Boolean question and 3 attempts for all others. Our final dataset included 409 distinct questions with 265,510 submission attempts overall. The submission time was recorded as well as the student's section. The offline test was completed on paper as part of the students' class session and includes multiple open-ended questions. The test was graded manually. Homework and test scores are floating numbers between 0 and 100, inclusive.

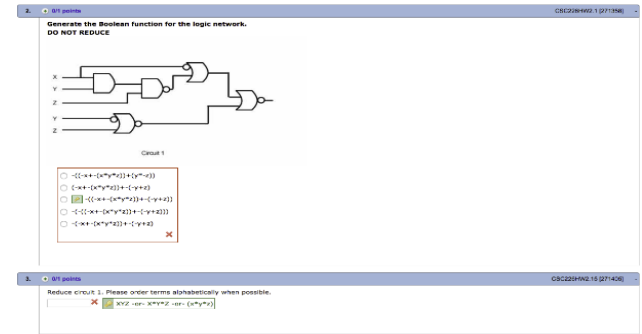

**Figure 2: A sample question on Webassign**

## 3. METHODS

We represented student performance with the features to represent shown in Table 1.

### 3.1 Feature Selection

We used the VarianceThreshold method from the Sci-Kit Learn Python library (version 0.19.0) to perform feature selection. The dataset includes some easy questions that almost every student answered correctly in one submission, indicating the corresponding features did not have much variance. Therefore, no good predictions can be made from these features, so we eliminated them from analysis. To save computing power and avoid spurious correlation between these features, we tested several combinations of thresholds of variance. The combinations tested for Per-Question Performance and Submissions Per Question respectively were (0.00, 0.00), (0.02,0.05), (0.03, 0.07), (0.04, 0.10). By checking the final accuracy of predictors after running the models under different thresholds, we found (0.02, 0.05) achieved the best accuracy. Therefore, we chose (0.02, 0.05) as the threshold in our analysis, and it selects 311 and 329 features for Per-Question Performance and Submissions Per Question, where original number of features are all 409.

**Table 1: The feature list**

| Feature | Total |
| --- | --- |
| **Per-Question Performance:** Whether a student answers questions correctly indicating skill mastery. | 409 |
| **Submissions Per Question:** Number of tries per question indicating the number of errors or guesses. | 409 |
| **Response Time:** Extra-long response times indicate that the student may be distracted or having difficulty with the question while very short response times may indicate guesses. Long responses are defined as response times two standard deviations above average while quick response times are > 5 per minute. We exclude responses that are longer than 2 hours as this indicates a disconnected session. | 4 |
| **Sessions Per Assignment**: A session is defined as a period of time taken on homework. Two adjacent tries within 2 hours are treated as the same session. Multiple sessions per assignment may indicate difficulty with the assignment. | 4 |
| **Homework and Test Scores:** are used in calculating the final grade. | 5 |

## 3.2 Normalization and Manual Segmentation

After eliminating uninformative features, we normalized each value to the range [0,1] to prevent any one feature from dominating the others. We then compared the performance of our trained models on both the normalized and unnormalized data to assess the impact of this step.

We also plotted the distribution of the submission attempts and response times for each question in order to assess their utility. Both distributions are dramatically right-skewed. Therefore, we did not expect manual segmentation from the decision tree to be more meaningful than the automatic segmentation provided by the Sci-Kit library. Therefore, we therefore opted not to perform any manual segmentation in this study.

## 3.3 Machine Learning and Cross Validation

We used the following standard implementations of the machine learning methods from the Sci-Kit library to train our models: Support Vector Machine (RBF kernel), Support Vector Regression, Decision Tree (Scikit-Learn uses an optimized version of the CART algorithm.), Naive Bayes and K-Nearest Neighbors (K=5). In order to assess the performance of the trained models, we also added two baseline models: random prediction and predicting the most frequent grade (A in this case).

We then estimated the stability of the models using Leave-One-Out cross validation we report the overall accuracy and a confusion matrix for each algorithm along with an, average precision score (micro over cross-validation), AUROC (exactly correct vs. not exactly correct), f1 score and mean squared error is also calculated to better compare the performances of models.

## 4. RESULTS

Because Support Vector Machine and Support Vector Regression use regularization (C=1.0) to prevent overfit, and the other models are sensitive to changes in attributes' values, the normalization process impacted performance. However, it did not lead to any consistent improvement in the accuracy relative to the non-normalized models. Because the accuracy of the Support Vector Machine and Linear Regression methods dropped significantly after normalization, we will focus solely on the non-normalized models in the remainder of the paper.

Based on Leave-One-Out cross validation, Support Vector Machines perform best among all the five algorithms, achieving 54.1% accuracy. Support Vector Regression, Decision Tree and K-Nearest neighbor reached more than 40% accuracy, but Naive Bayes performed just slightly above chance. The other performance statistics showed the same trend.

**Table 2: Performance for non-normalized input**

|  | Accuracy | Mean Square Error | Average Precision (Micro) | AUROC | f1 score |
|---|---|---|---|---|---|
| SVM | 51.4% | 1.755 | 0.36 | 0.573 | 0.514 |
| Lin. Reg | 45.8% | 1.304 | 0.23 | 0.558 | 0.289 |
| Decision Tree | 43.8% | 1.803 | 0.3 | 0.590 | 0.437 |
| Naive Bayes | 24.1% | 3.108 | 0.21 | 0.539 | 0.240 |
| KNN | 41.8% | 1.510 | 0.29 | 0.595 | 0.417 |
| Random | 20.0% | 4.807 | 0.2 | 0.492 | 0.196 |
| All A | 47.8% | 2.674 | 0.33 | 0.500 | 0.477 |

**Table 3: Performance for normalized input**

|  | Accuracy | Mean Square Error | Average Precision (Micro) | AUROC | f1 |
|---|---|---|---|---|---|
| SVM | 51.0% | 2.160 | 0.36 | 0.536 | 0.510 |
| Lin. Reg | 23.7% | 1.459 | 0.23 | 0.523 | 0.301 |
| Decision Tree | 42.2% | 1.702 | 0.29 | 0.576 | 0.421 |
| Naive Bayes | 25.3% | 3.108 | 0.21 | 0.543 | 0.253 |
| KNN | 24.1% | 1.767 | 0.21 | 0.554 | 0.240 |
| Random | 20.0% | 4.807 | 0.2 | 0.492 | 0.196 |
| All A | 47.8% | 2.674 | 0.33 | 0.500 | 0.477 |

We then generated confusion matrices for the different approaches. These matrices are shown in Tables 4 & 5. Here the difference is the absolute distance between the predicted grade and the actual grade on an integer scale (5-A 4-B 3-C 2-D 1-F)

**Table 4: Confusion matrix for unnormalized input**

|  | 0 | 1 | 2 | 3 | 4 |
|---|---|---|---|---|---|
| SVM | 128 | 81 | 19 | 8 | 13 |
| Lin. Reg | 114 | 89 | 34 | 7 | 5 |
| Decision Tree | 109 | 94 | 23 | 15 | 8 |
| Naive Bayes | 60 | 82 | 78 | 12 | 17 |
| KNN | 104 | 98 | 36 | 6 | 5 |
| All predict to A | 119 | 72 | 22 | 10 | 26 |

**Table 5: Confusion matrix for normalized input**

| Difference | 0 | 1 | 2 | 3 | 4 |
|---|---|---|---|---|---|
| SVM | 127 | 72 | 22 | 10 | 18 |
| Lin. Reg | 59 | 159 | 13 | 16 | 2 |
| Decision Tree | 105 | 98 | 26 | 14 | 6 |
| Naive Bayes | 63 | 78 | 79 | 12 | 17 |
| KNN | 60 | 116 | 68 | 4 | 1 |

## 5. DISCUSSION & FUTURE WORK

While the machine learning models described in this study can be used to predict students' final grades to some extent, the accuracy is still far from ideal for real-world applications. Although the best model (SVMs) performed better than the naive baseline models, the advantage is not significant. At the same time, normalization did not bring us any notable improvement. When examining the misclassified students, we found that a considerable portion of students who did well in the homework actually performed poorly in the first test. Given the high percentage (47.8%) of A grades in this course and the fact that homework typically permitted multiple tries we concluded that the homework may have been too easy, and that students' final homework scores were not reliable predictors of their future test scores, which are in turn the largest portion of final score. Thus, it was not possible to derive a good predictive model that relies heavily on homework submission logs. We also noticed that almost all of the students who did not complete or performed poorly in one of the assignments eventually dropped the course. We believe that these are students who may have wanted to drop the course and who thus quit doing the homework before dropping or who were motivated to do so after a particularly bad homework score. Unfortunately, none of the models correctly captured this phenomenon.

In the future, we hope to examine if feature engineering can be used to address the limitations above. If we can predict dropouts in advance, then we can make the models much more robust. One other possible way to improve upon this is to add additional features. A richer model may be more robust in the face of noise. Combining this interaction model with models based on social network data, for example, may improve our performance particularly in cases where help-seeking is an important indicator of performance. Brown et. al [4] have shown that students on MOOCs formed detectable communities, and community membership was significantly correlated with performance. In addition, Gitinabard et. al [6] showed that students who asked more questions and received more feedback on the forum tended to obtain higher grades in blended courses. It will be interesting to see if students closely connected in a social network in course influence each other and further change the homework pattern of features overtime.

## ACKNOWLEDGMENTS

The authors wish to thank Zhongxiu Aurora Liu and members of that Center of Educational Informatics at North Carolina State University for their assistance.

This research was partially supported by the National Science Foundation Grant #1418269: "Modeling Social Interaction & Performance in STEM Learning" Yoav Bergner, Ryan Baker, Danielle S. McNamera, & Tiffany Barnes Co-PIs.